\documentclass[letterpaper, 10pt, conference]{ieeeconf}                             
\IEEEoverridecommandlockouts         
                                                     
\usepackage{times} 
\usepackage{amsmath}
\usepackage{amssymb,longtable,calc}
\usepackage{mathptmx}
\usepackage[T1]{fontenc}                                                        
\usepackage[utf8]{inputenc}                                                     
\usepackage[english]{babel}                                                     
\usepackage{epsfig}                                                             
\usepackage{subfigure}                                                          
\usepackage{textcomp} 
\usepackage[textwidth=2cm,colorinlistoftodos]{todonotes} 
\usepackage{tikz}                                                               
\usetikzlibrary{arrows,positioning,fit,shapes,calc}
\usetikzlibrary{matrix}
\usepackage{flushend}                                                           
\usepackage{hyperref}  
\usepackage{amsmath}    

\usepackage[utf8]{inputenc}   
\usepackage[]{algorithm2e}
\usepackage{amsmath}

\usepackage{pgfplots} 
\usepackage{pgfplotstable}

\usepackage{cite}

\usepackage[tikz]{bclogo}
\usepackage{lipsum}

\usepackage{standalone}

\pgfplotsset{compat=newest}
\pgfplotsset{ 
  tick label style={font=\footnotesize}, 
  label style={font=\footnotesize}, 
  legend style={font=\footnotesize},
  title style = {font=\small}
}
\pgfplotscreateplotcyclelist{line style}{%
  solid, mark options = {scale = .75}, every mark/.append style={fill=gray},mark=*\\%
  densely dashed,mark options = {scale = .75},every mark/.append style={solid,fill=gray},mark=*\\%
  densely dotted,mark options = {scale = .75},every mark/.append style={solid,fill=gray},mark=*\\%
  dashed,mark options = {scale = .75},every mark/.append style={solid,fill=gray},mark=*\\%
  dotted,mark options = {scale = .75},every mark/.append style={solid,fill=gray},mark=*\\%
}
\pgfplotscreateplotcyclelist{bar style}{%
  solid, fill=black!60!white\\%
  solid, fill=black!45!white\\%
  solid, fill=black!35!white\\%
  solid, fill=black!25!white\\%
}

\usepackage{xspace}
\makeatletter                                                                   
\DeclareTextCommandDefault{\textregisteredalt}{\footnotesize\textcircled{%
      \check@mathfonts\fontsize\sf@size\z@\math@fontsfalse\selectfont R}}       
\DeclareRobustCommand\onedot{\futurelet\@let@token\@onedot}                     
\def\@onedot{\ifx\@let@token.\else.\null\fi\xspace}

\makeatother
                                                                    
\definecolor{lightGray}{rgb}{0.0,0.0,0.0}
\definecolor{kthColor}{RGB}{26,84,166}
                                                                                
\title{\LARGE \bf Feature Descriptors for Tracking by Detection: a Benchmark}

\author{Alessandro Pieropan ~~~~ Mårten Bj{\"o}rkman  ~~~~ Niklas Bergstr{\"o}m ~~~~ Danica Kragic%
\thanks{This research has been supported by he Japan Society for the Promotion of Science (JSPS)}
\thanks{The GPU used for this research was donated by the NVIDIA Corporation.}
\thanks{MB and DN are with CVAP/CAS, KTH, Stockholm, Sweden, {\tt celle,dani@kth.se}. AP and NB are with the University of Tokyo, Japan, {\tt alessandro\_pieropan}, {\tt niklas\_bergstrom@ipc.i.u$-$tokyo.ac.jp}.}}

\begin{document}                                                                
                                                                                
\maketitle                                                                      
\thispagestyle{empty}                                                           
\pagestyle{empty}

\begin{abstract}
In this paper, we provide an extensive evaluation of the performance of local descriptors for tracking applications.
Many different descriptors have been proposed in the literature for a wide range of application in computer vision such as object recognition and 3D reconstruction. More recently, due to fast key-point detectors, local image features can be used in online tracking frameworks. However, while much effort has been spent on evaluating their performance in terms of distinctiveness and robustness to image transformations, very little has been done in the contest of tracking. Our evaluation is performed in terms of distinctiveness, tracking precision and tracking speed. Our results show that binary descriptors like ORB or BRISK have comparable results to SIFT or AKAZE due to a higher number of key-points.    

\end{abstract}

\section{INTRODUCTION}
\label{sec:introduction}

Local regions of interest or key-point descriptors are widely used in Computer Vision for application such as object recognition and retrieval, 3D reconstruction and motion tracking. SIFT \cite{lowe04} is widely considered as one of the most robust feature descriptors, providing distinctiveness and invariance to common image transformations such as rotation and scale. However such a robustness comes at a considerable computational cost. 

Recently there is much concern about efficiency caused mainly by two important factors. First there is a stable growth of portable camera enabled devices with limited computing power. Second, databases used for computer vision application are steadily increasing in size. As a consequence, there is a growing interest within the computer vision community it fast key-point detectors and binary descriptors that can dramatically decrease the computational cost of detecting and matching local regions of interest. BRIEF \cite{calonder10} feature descriptor in combination with FAST \cite{rosten06} key-point detector is among the first attempts in this direction making it suitable for real time applications. However, despite the improvement in performance, BRIEF descriptor is not very robust to image transformations. This underlines the difficulty in finding a good compromise between two competing characteristics: distinctiveness and fast computation. 

\begin{figure}[t]
	\vspace{2mm}
\centerline{%
	\subfigure{\includegraphics[width=0.48\linewidth]{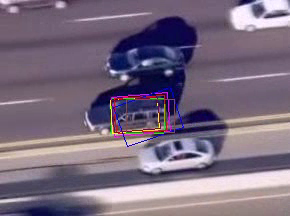}}
	\subfigure{\includegraphics[width=0.48\linewidth]{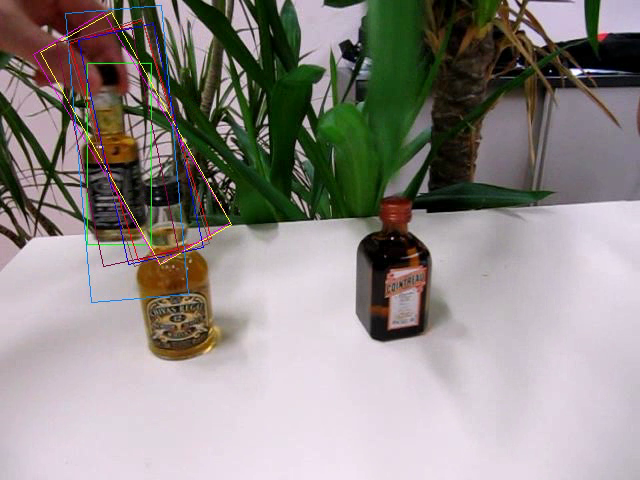}}}
	\vspace{-2mm}
\centerline{%
	\subfigure{\includegraphics[width=0.48\linewidth]{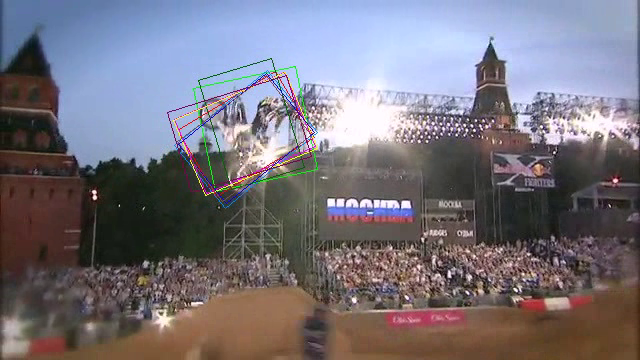}}
	\subfigure{\includegraphics[width=0.48\linewidth]{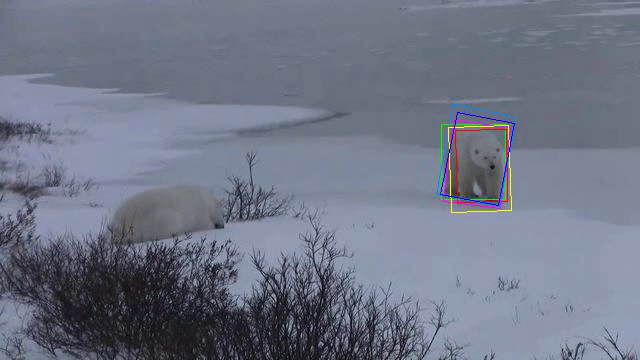}}}
\caption{Examples showing some of the videos the descriptors has been tested on and the tracking results expressed as a coloured bounding box. Clearly some feature descriptors show more precise tracking.}
\vspace{-3mm}
\label{fig:intro}
\end{figure}

More attempts has been done in this direction. BRISK \cite{leutenegger11} and ORB \cite{rublee11} include some modification of the BRIEF and FAST in order to achieve scale and rotation invariance. The mentioned descriptors are faster than SIFT and have comparable matching precision in case of small image transformation. More recently the binary feature descriptor KAZE \cite{alcantarilla12} had comparable results to SIFT on a standard dataset \cite{mikolajczyk05} designed to evaluate the robustness of local descriptors on several image transformations. Moreover, by using fast non-linear filtering techniques Fast Explicit Diffusion \cite{goesele2010}, its accelerated version AKAZE \cite{alcantarilla13} has shown to be competitive to SIFT in computational cost.\\
The main reason why AKAZE could outperform SIFT relies in the use of non-linear filtering techniques. However, given the increasing use of GPUs in computer vision, it is not clear if such techniques can be easily parallelizable on a GPU architecture as opposed to Gaussian filtering employed in SIFT. A recent work \cite{jiang2015} has shown that it is possible to deploy AKAZE descriptor on a specialized hardware and achieve a good speedup compared to the original implementation. By implementing the descriptor in CUDA we intend to analyze its performance using a GPU and compare it against the GPU implementations of SIFT descriptor.


These fast local descriptors not only improve the performance of vision tasks such as object retrieval and 3D reconstruction but they can be used in real-time tracking system. Yet, to the best of our knowledge, very little work has been done in evaluating key-points descriptor for the specific purpose of tracking. This work aims to target that issue by providing a fair and comprehensive evaluation of the most well known descriptors. We use the recall-precision measure to evaluate the matching distinctiveness of the features and we calculate the tracking precision by integrating each local descriptor in a key-point based tracker \cite{pieropan15}. Since there is no dataset designed to test local descriptors for tracking purposes, the videos used in the evaluation are gathered from well known datasets used to evaluate tracking algorithms in general, described in \cite{wu2013,nebehay2014,hare2011}. We want to provide a practical guideline to descriptors for the specific task of tracking since it is not clear yet that a descriptor designed for image recognition is well suited for tracking.
 
The contribution of this work is four-fold:\\
\textbf{Dataset.} We collected 47 sequences from different public available datasets and annotated the ground-truth in a standardized format to ease the evaluation.\\
\textbf{Cuda AKAZE.} Given the growing interest in the AKAZE feature descriptor and the lack of a GPU implementation, we provide our own using CUDA.\\
\textbf{Evaluation Library.} We integrated the most well known descriptors in a state-of-the-art key-point based 2D tracker and we provide an interface to enable the integration of new feature descriptors.\\
\textbf{Evaluation.}  There are three criteria the local descriptors are tested upon: distinctiveness, tracking precision and speed. First we measure the distinctiveness by matching the feature descriptors extracted in the first frame of the sequence and by computing the recall-precision. Then we calculate the tracking by integrating each feature descriptor in a key-point based tracker and we calculate the well known overlap accuracy for low, medium and high precision requirements. Last we profile the performance of each descriptor.

The dataset, our AKAZE implementation and all the code to perform the benchmark will be publicly available.

\section{Related Work}
\label{sec:relatedwork}

Evaluating the performance has a crucial role in computer vision in order to test the reliability of the algorithms and determine if they are of any practical use \cite{christensen02,butler12,wu2013}. Evaluating feature descriptors is not an exception. One of the most relevant work has been done in 2005 by Mikolajczyk et al. \cite{mikolajczyk05} where affine region detectors are evaluated in terms of repatability and accuracy. SIFT feature descriptor \cite{lowe04} and its extension GLOH \cite{mikolajczyk05} had the best results on the dataset decribed in \cite{mikolajczyk2005b}. In recent years, due to a steady increase of portable devices and large scale databases, there is more attention to the runtime requirements of such descriptors. SURF \cite{bay2008} is among the first descriptors that presented lower computational requirements than SIFT and comparable precision. More recently, Calonder et al. \cite{calonder10} proposed to use a binary descriptor BRIEF, as opposed to gradient histograms used in SIFT, in combination with a fast corner detector FAST \cite{rosten06} achieving a fraction of the runtime performance of SURF. However, the descriptor has comparable results only when small transformations are applied to an image since the it is not scale or rotation invariant. ORB descriptor (Oriented Fast and Rotated Brief) proposed by Rublee et al. \cite{rublee11} enhances BRIEF adding the rotation invariance. Another binary descriptor, BRISK \cite{leutenegger11}, provides both scale and orientation invariance. Its corner detector, AGAST \cite{mair2010} , is also faster than FAST however its invariance properties influence the overall performance.\\ 
Heinly et al. \cite{heinly2012} evaluated exhaustively these three binary descriptors using SURF and SIFT as baseline on the Oxford dataset \cite{mikolajczyk2005b}. Not surprisingly, SIFT was the most precise while BRIEF was the fastest one but, since it is not rotation and scale invariant, its performance is significantly lower than the other descriptors upon large image transformation. This evidence underlines the challenge in finding a good compromise between precision and speed. Nevertheless, the small computational requirements of such descriptors drawn the attention on their potential use in real-time application such as tracking or simultaneous mapping and localization (SLAM). \\
A promising binary feature, KAZE, has been proposed by Alcantarilla et al. \cite{alcantarilla12}. Its main strength relies on using non-linear diffusion filtering techniques \cite{weickert98} to build the scale space as opposed to Gaussian blurring. Since the latter smooths without distinction at any scale, it does not preserve object boundaries. On the other hand non-linear diffusion filtering techniques preserve edges, resulting in an increased distinctiveness of the feature descriptor. Once again, an improved precision comes at a cost in speed, however by the use of Fast Explicit Diffusion \cite{goesele2010} the accelerated version of the feature descriptor AKAZE \cite{alcantarilla13} the descriptor could achieve better accuracy than SIFT. Yet the experiments has been performed on the Oxford dataset, designed to evaluated the precision of feature descriptors with image transformation. Such features are now employed in real-time system such as tracking by detection \cite{nebehay2014,pieropan15,pieropan15b} or more in general SLAM \cite{murartal2015}.


\section{Benchmark}

The goal of the conducted benchmarking presented in detail in the experimental evaluation is three-fold. First, we want to measure the descriptiveness of the feature descriptors for matching purposes. This is crucial for the recovery of a tracker upon loss of tracking due to occlusion, changed lighting conditions, and alike. Second, we measure the tracking accuracy by integrating each feature descriptor in our 2D tracker and compute the overlap measure using the estimated object position. Third, we profile each separate step required in tracking by detection (key-point detection, descriptor computation and feature matching) in order to evaluate the performance of the feature descriptors for real-time applications.

\subsection{Dataset}

\begin{figure*}[t]
	\vspace{2mm}
\centerline{%
	\subfigure{\includegraphics[width=0.19\linewidth]{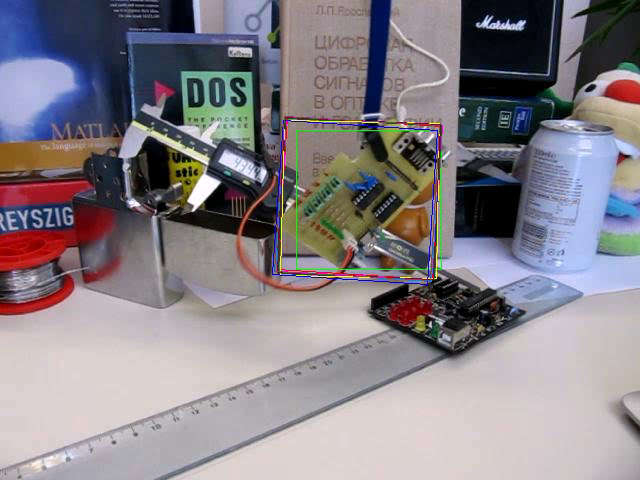}}
	\subfigure{\includegraphics[width=0.19\linewidth]{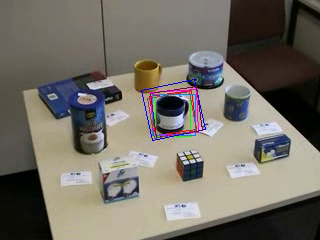}}
	\subfigure{\includegraphics[width=0.19\linewidth]{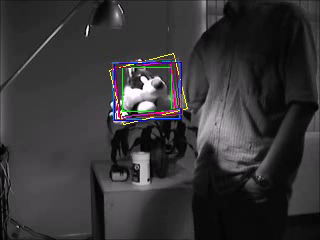}}
	\subfigure{\includegraphics[width=0.19\linewidth]{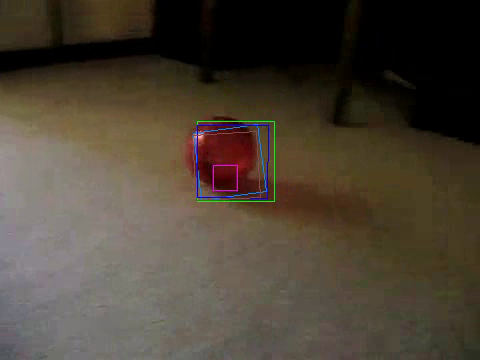}}
	\subfigure{\includegraphics[width=0.19\linewidth]{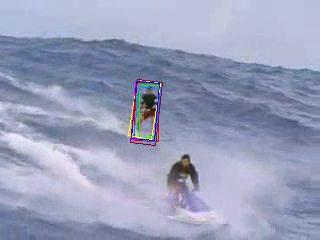}}}
	\vspace{-2mm}
\centerline{%
	\subfigure{\includegraphics[width=0.19\linewidth]{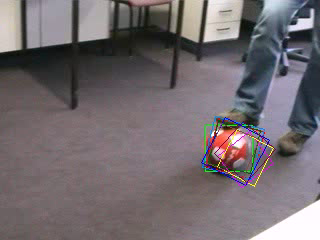}}
	\subfigure{\includegraphics[width=0.19\linewidth]{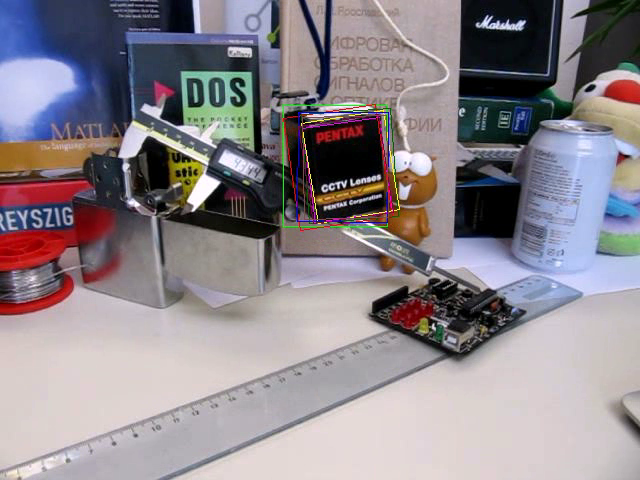}}
	\subfigure{\includegraphics[width=0.19\linewidth]{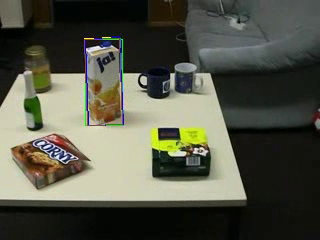}}
	\subfigure{\includegraphics[width=0.19\linewidth]{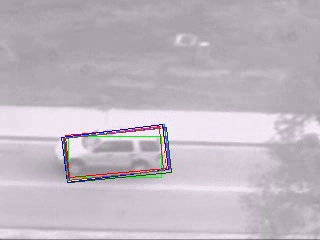}}
	\subfigure{\includegraphics[width=0.19\linewidth]{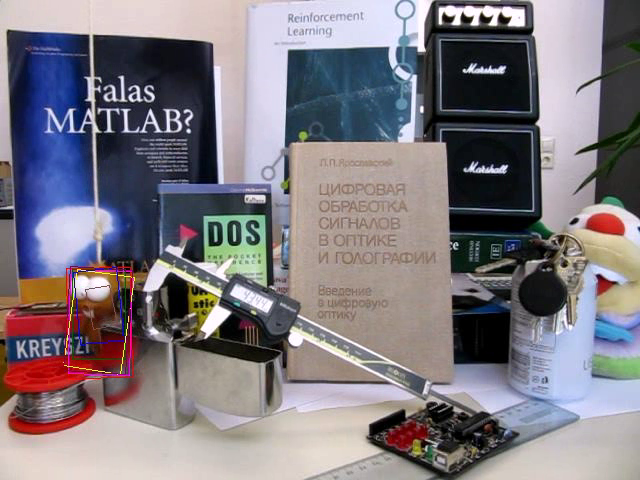}}}
\caption{A few example images from the videos used in the benchmark. The sequences include both indoor and outdoor scenes, subjected to changing lightning conditions, 
occlusion, and significant changes in object appearance.}
\vspace{-3mm}
\label{fig:tracking_results}
\end{figure*}

There are several publicly available datasets designed for tracking. However, there is no agreement in the community of the standard structure to store the data. For this reason and in order to facilitate the evaluation, we collected the videos from the different datasets and standardized how these are stored. Each video is stored as a sequence of images while the ground truth, represented by an oriented bounding box of the tracked object, is saved in a separate file where each row corresponds to an image frame defined by the 8 values representing the pairs (x,y) of each vertex of the frame. The generated dataset will be released publicly, along with all the code to perform the experiments and run the evaluation.

\subsection{Matching Evaluation Criteria}
Given a sequence of images $I_{1},...,I_{n}$ with corresponding ground truth bounding boxes $B^g_{1},...,B^g_{n}$ around a target object, we extract a set of key-points $K_1$ from the first image of the sequence and regard all key-points within $B^g_{1}$ as descriptors of the object. For each subsequent image key-points $K_{t}$ are similarly extracted and then matched to those in the initial set $K_{1}$, generating a list of matches $M_t = \left\lbrace (i,j); ~p_{1,i} \in K_1, p_{t,j} \in K_t \right\rbrace$. Given the bounding boxes a key-point $p_{t,j}$, with a match $(i,j)\in M_t$, is then labeled as follows:
\begin{equation}
\begin{cases}
\text{true positive},&  \text{ if } p_{t,j} \in B^g_{t} \land p_{1,i} \in B^g_{1} \\
\text{false positive},&  \text{ if } p_{t,j} \notin B^g_{t} \land p_{1,i} \in B^g_{1} \\
\text{false negative},&  \text{ if } p_{t,j} \in B^g_{t} \land p_{1,i} \notin B^g_{1} \\
\end{cases}
\end{equation}



\begin{figure}[!htb]
	\includegraphics[width=0.95\linewidth]{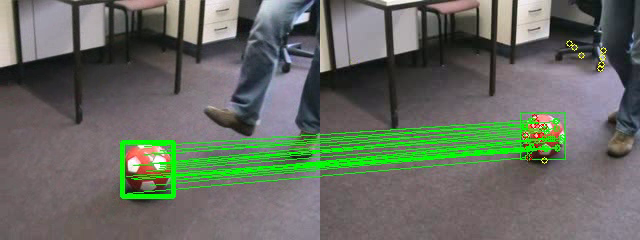}
\vspace{-2.5mm}	
\caption{Example showing how the matching precision is calculated. The image to the right shows the initial frame of the sequence. Matches in green are true positives. Circles in red are false positives. Yellow circles are false negatives, feature descriptors that are inside the object in the current frame but have a match with the background in our model.}
\label{fig:matching}
\end{figure}

The average ratio of true positives assesses the ability of a tracker to detect the object and potentially recover after the loss of tracking. False negatives are those feature descriptors that appear in the current frame but have no corresponding match to the initial test set $K_{1}$. This may commonly happen when there is a drastic change in appearance of the object. The ratio of false positives is very important to consider since it indicates the average number of outliers that will be used to estimate the pose of the object, resulting in a bad pose estimation if no additional filtering techniques are employed. A widely used filtering technique is to discard all matched key-points if the ratio between the score of the best match and the second best match is below a certain threshold $\rho$. We define a key-point as ambiguous if this criteria is not met. The  number of ambiguous true and false positives is then calculated to evaluate the distinctiveness of the descriptors and evaluate the influence of this common filtering technique on the results.

\subsection{Evaluating tracking precision}
\label{sec:accuracy}

We employ a sparse key-point based tracker to measure the precision of the feature descriptors for tracking. The tracker is initialised using a bounding box in the initial image of a tracking sequence. Inside the bounding box, feature descriptors are extracted and represent the \textit{model} of the tracked object. The algorithm estimates the position of the object, represented as an oriented bounding box, with a combination of sparse optical flow and feature matching as outlined in Algorithm \ref{alg:algorithm}. For the complete details please refer to \cite{pieropan15}.

\begin{algorithm}[!htb]
 \KwData{$I_{1},...,I_{n},B_{1}$}
 \KwResult{$B_{2},...,B_{n}$}
 $K_{1} \gets$ \textbf{extract\_points}($I_{1},B_1$)\;
 \For{$i \gets 2 : n$}{
   $K_{i}^{*} \gets$ track\_points($K_{i-1},I_{i-1},I_i$)\;
   $B_i \gets$ estimate\_pose($K_{i}^{*}$)\;
   $K_{i}' \gets$ \textbf{extract\_points}($I_{i},B_i$)\;
   $M \gets$ \textbf{match\_points}($K_{1},K_{i}'$)\;
   $K_{i} \gets$ merge\_keypoints($K_i^* ,  M$)\;
 }
 \caption{\label{alg:algorithm}Overview of the tracking algorithm used to compute the tracking precision. The feature descriptors are employed in the steps written in bold. }
\end{algorithm}

Our original tracking algorithm used ORB (Oriented Fast and Rotated Brief) proposed by Rublee et al.~\cite{rublee11}. 
We extended the tracking algorithm making it more modular so that tracking can be performed with 
various feature descriptors that we evaluate. To estimate the tracking precision, we used the widely accepted overlap measure:
\begin{equation}
	\Theta (B_{t}, B^g_t) = \frac{B_{t} \cap B^g_t}{B_{t} \cup B^g_t}
\end{equation}
where \textit{$B_{t}$} is the bounding box estimated by our tracker and
\textit{$B^g_t$} is the bounding box provided by the ground truth. We define 
three precision requirements $\Upsilon \in \{0.25, 0.5, 0.75\}$ that indicate low, medium and high tracking accuracy. This is a more indicative evaluation compared to the overall accuracy. For instance, an overall value of 0.5 is ambiguous because it may indicate either a stable average accuracy around the value or a very precise evaluation in part of the video while poor in the rest.
%
%
%
%
%
The overall accuracy of the tracker is then computed for each precision requirement $\Upsilon$ as the fraction of all generated bounding boxes for which $\Theta(B_{t}, B^g_t) > \Upsilon$.


\subsection{Parameter setting}

All the employed descriptors require many parameters to be initialised and run properly. In order to achieve the most fair comparison between them, we decided to keep the values suggested in the original publication of the feature descriptor or the implementation if such was available. However, there are some parameters that most of the feature descriptors share and that we set to the same value. Thus, the maximum number of features extracted is set to 2500 and the number of scale space octaves is set to 4.

\section{Results}

In this section, we summarise the results of our assessment. The descriptors considered in the evaluation are: BRISK, ORB, SURF and SIFT included in OpenCV library. The implementation of AKAZE made available by its original author \cite{alcantarilla13} and the implementation of SIFT using CUDA presented in \cite{bjorkmann14}.

\subsection{Distinctiveness}


The first aspect was to assess the distinctiveness. Table~\ref{table:tp_ratio} shows the average number of key-points extracted, descriptors belonging to the object to track and the ratio of true positives (TP), false positives (FP) and true positives that pass the second best match ratio check (TTP). 

\begin{table}[!h]
\caption{Average number of feature extracted, object features, true positives and false positives. Every row is normalized by its maximum value.}
\vspace{-2mm} 
\centerline{%
		\includegraphics[width=0.98\linewidth]{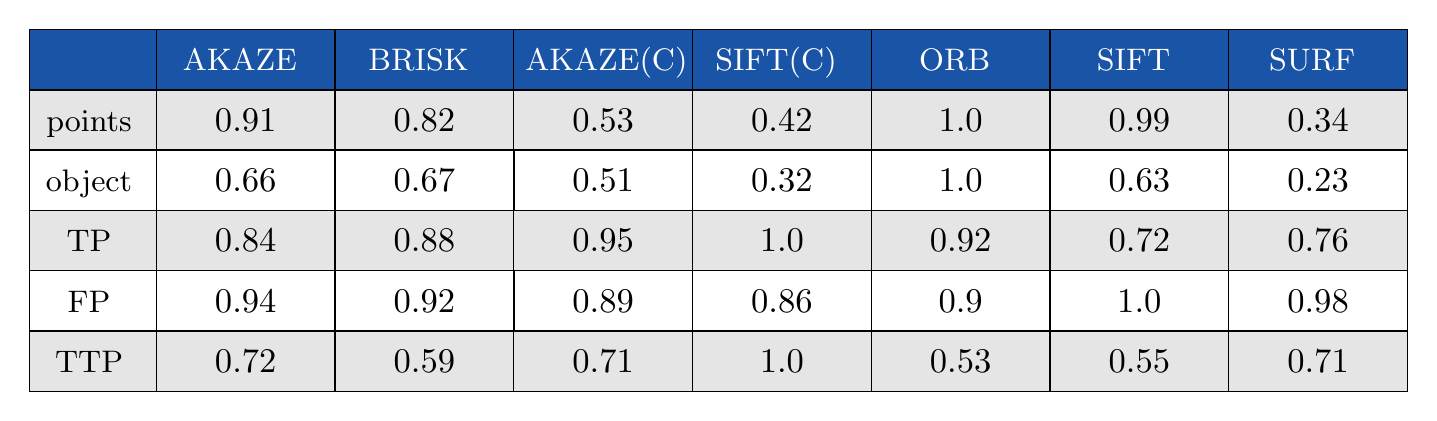}}
    \vspace{-2mm} 
	\label{table:tp_ratio}
\end{table}

It is interesting to notice that BRISK, ORB and SIFT extract a higher number of feature descriptors in general. In particular, BRISK and ORB have a higher number of key-points extracted within the area of the object. However, looking at the average amount of true positives, it can be seen that the best performing descriptors are AKAZE and the implementation of SIFT on the GPU. This is a first indicator of the quality of the descriptors extracted. Moreover, it can be noticed that the true positives are also more distinctive in the case of AKAZE and SIFT since the number of TTP is higher.

\begin{table*}[t]
\caption{Tracking results with low, medium and high accuracy requirements. The high number of key points extracted by ORB or BRISK compensate their weak descriptors. This comes with a cost in performance.} 
\centerline{%
		\includegraphics[width=\linewidth]{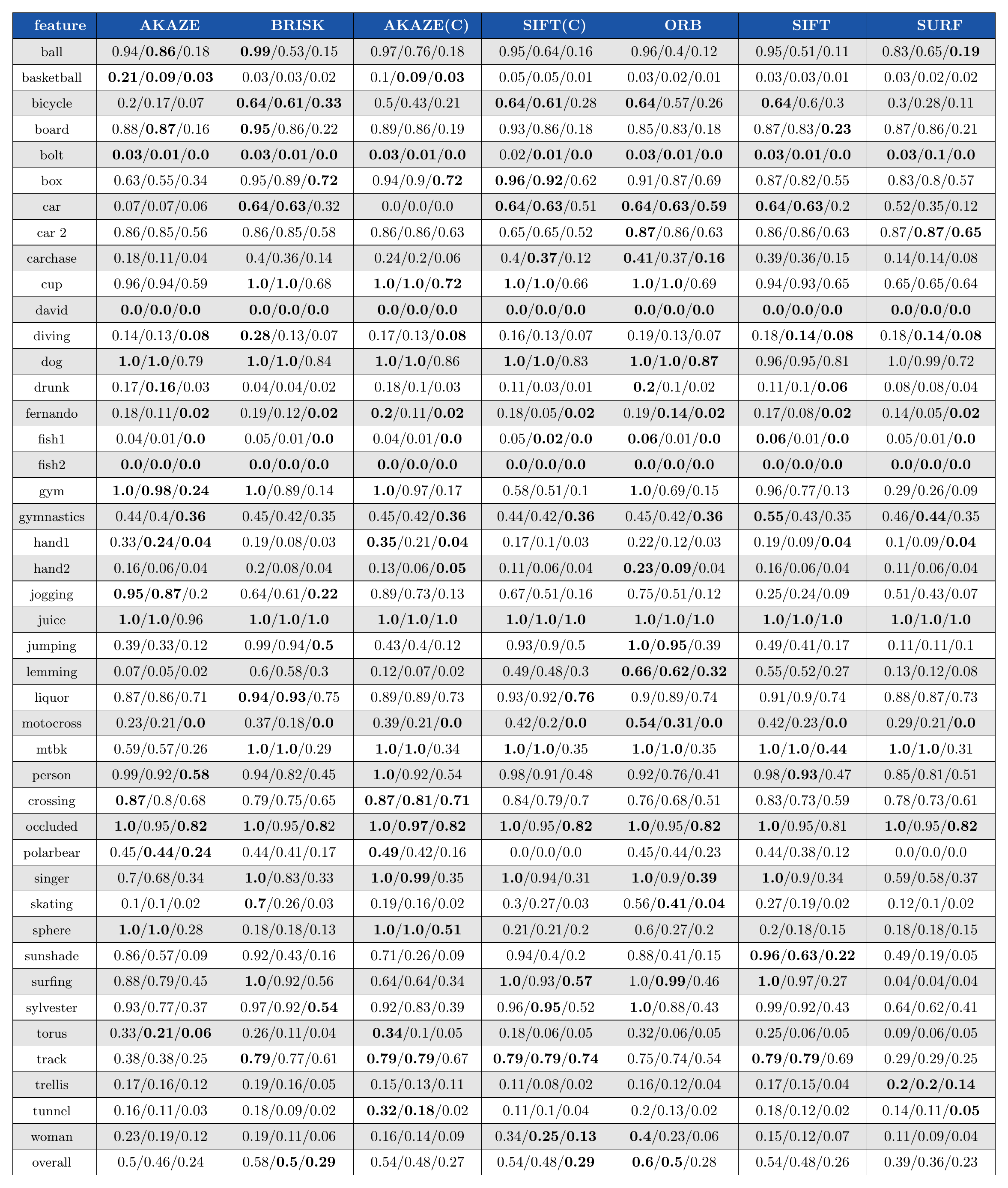}}
		\vspace{8mm}
	\label{table:taccuracy}
\end{table*}

\subsection{Tracking accuracy}

As explained in the previous section, we evaluated the accuracy of the feature descriptors by running our tracker and calculating the overlap measure for low, medium and high accuracy requirements. 
\begin{figure}[!h]
	\vspace{2mm}
\centerline{%
	\subfigure{\includegraphics[width=0.48\linewidth]{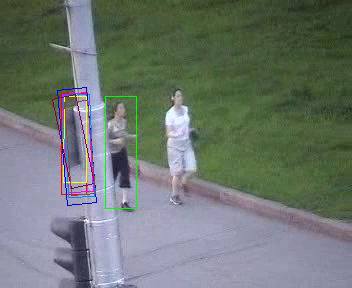}}
	\subfigure{\includegraphics[width=0.48\linewidth]{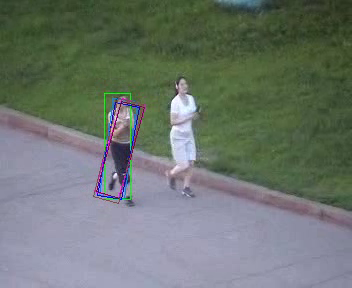}}}
\caption{Examples showing the behaviour of the feature descriptors upon occlusion. Upon recovery from track loss more descriptive descriptors allow the tracker to recover faster.}
\vspace{-3mm}
\label{fig:tracking_results}
\end{figure}

Table~\ref{table:taccuracy} summarizes tracking results on all the video sequences included in the dataset.  Our experiments show that AKAZE, BRISK, ORB and SIFT have comparable results. It is interesting to note that BRISK and ORB compensate their weak distinctiveness with a higher amount of weak descriptors extracted. A high number of feature points proved to be effective in tracking in the video sequences where the object suffers drastic scale changes and full occlusion, making the recovery after track loss faster, see Fig.~\ref{fig:tracking_results}. We also noticed that AKAZE, more than SIFT, suffers the change in scale. 

\subsection{Tracking performance}

\begin{figure}[!htb]
	\includegraphics[width=0.95\linewidth]{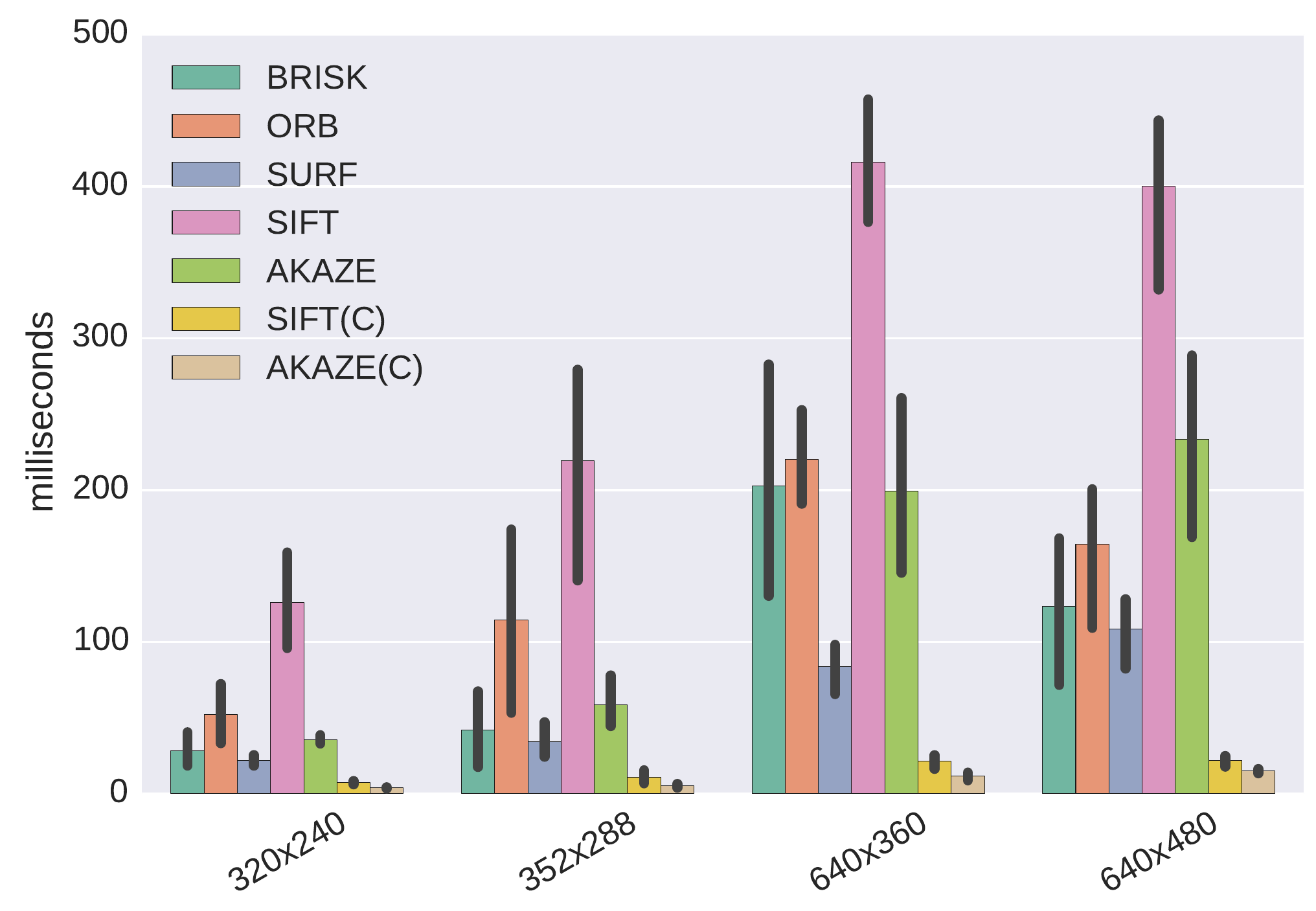}
\vspace{-2.5mm}	
\caption{Average time spent on tracking the object in a single frame: important factors are the resolution and the number of feature descriptors extracted. The variance of the results is a good indication of how much the number of feature descriptors influences the performance. It can be seen that the implementations have a lower variation due to the high level of parallelism. }
\vspace{-2mm}
\label{fig:speed}
\end{figure}

The dataset used for benchmarking includes video sequences of various resolution. Fig.~\ref{fig:speed} shows the average performance of each feature descriptor on the resolutions having the highest number of sequences. 

\begin{figure}[!htb]
	\includegraphics[width=0.95\linewidth]{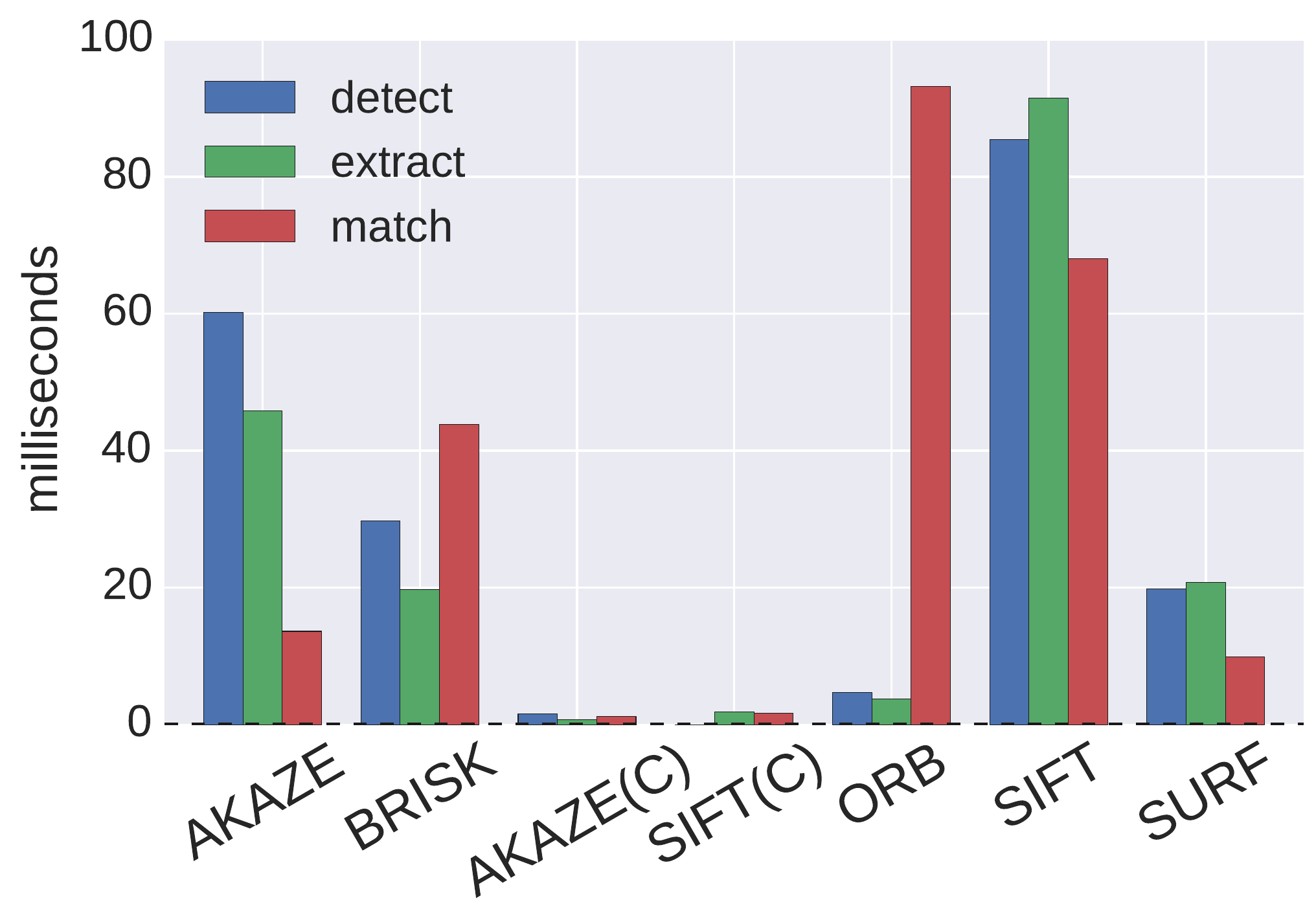}
\vspace{-2.5mm}	
\caption{Performance of the detect, extract and match steps of each feature descriptor.}
\label{fig:speed_b}
\end{figure}

The two most important factors that influence performance are resolution and number of key points extracted. The former influences particularly the detection step when the scale space of descriptors is computed and key-points are detected. The latter influences more the extraction step when feature descriptors are calculated and the matching step. The average performance of each separate step can be seen in Fig.~\ref{fig:speed_b}. One interesting aspect to notice is the variance of the performance in Fig.~\ref{fig:speed}: BRISK, ORB and SIFT have the higher variance while CUDA SIFT and CUDA AKAZE have lower variance. This is also a good indicator of the level of parallelism of the implementation of the descriptor. It is interesting to notice that the computation of the non linear scale space required by AKAZE is not perfectly suited for a GPU architecture since it requires many sequential steps, as a result the key-detector is slower that SIFT. However since the descriptor is binary, its extraction and matching compensate in terms of performance. It is also important to remember that CPU implementations do not exploit the same number of cores, complicating a fair comparison between methods.

\subsection{Discussion}

\begin{figure*}[!htb]
	\vspace{2mm}
\centerline{%
	\subfigure[Crossing: constant distance from camera, target often occluded.]{\includegraphics[width=0.24\linewidth]{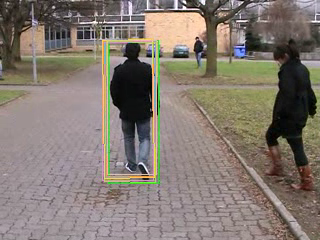}
			   \includegraphics[width=0.24\linewidth]{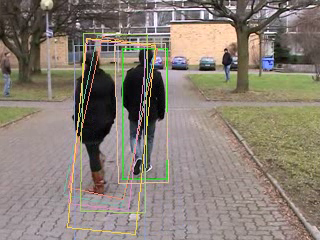}
			   \includegraphics[width=0.24\linewidth]{occ2.png}
			   \includegraphics[width=0.24\linewidth]{occ2.png}}
			   \label{fig:crossing}}
	\vspace{-2mm}
\centerline{%
	\subfigure[Jumping: high motion blur.]{\includegraphics[width=0.24\linewidth,trim={0, 0 0 2cm},clip]{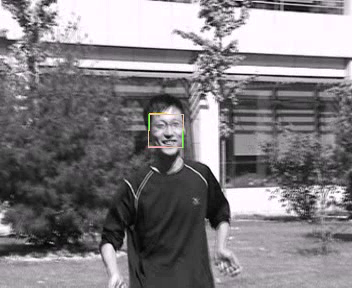}
	           \includegraphics[width=0.24\linewidth,trim={0, 0 0 2cm},clip]{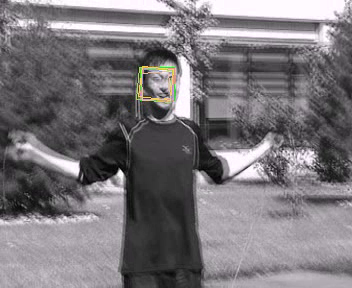}
			   \includegraphics[width=0.24\linewidth,trim={0, 0 0 2cm},clip]{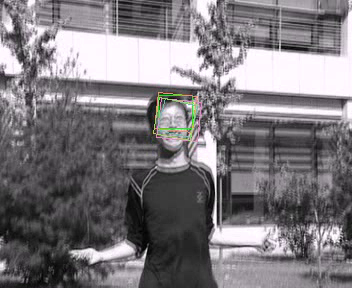}
	           \includegraphics[width=0.24\linewidth,trim={0, 0 0 2cm},clip]{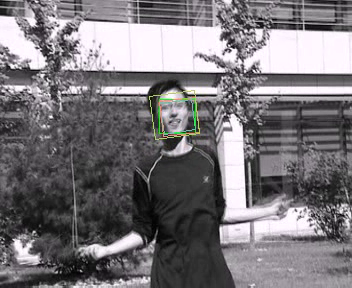}}
	           \label{fig:jumping}}
	           \centerline{%
\subfigure[Skating: motion blur, scale and light change.]{\includegraphics[width=0.24\linewidth]{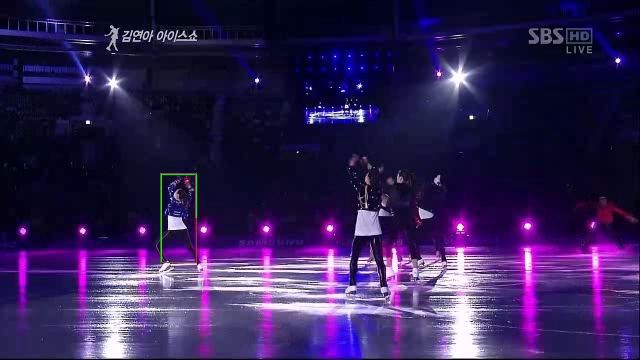}
	           \includegraphics[width=0.24\linewidth]{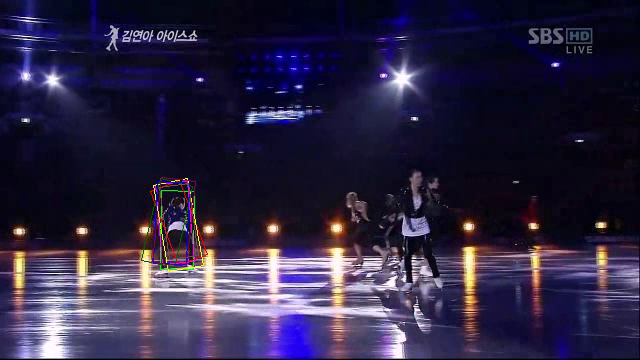}
	           \includegraphics[width=0.24\linewidth]{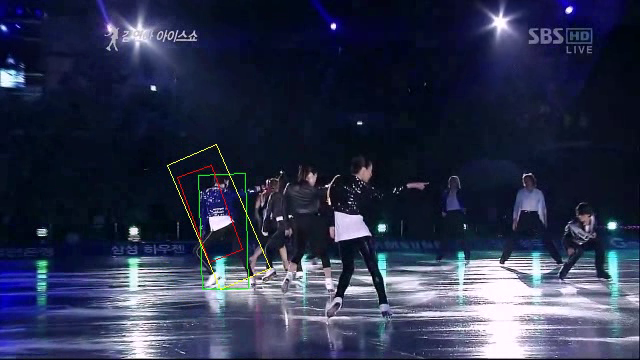}
	           \includegraphics[width=0.24\linewidth]{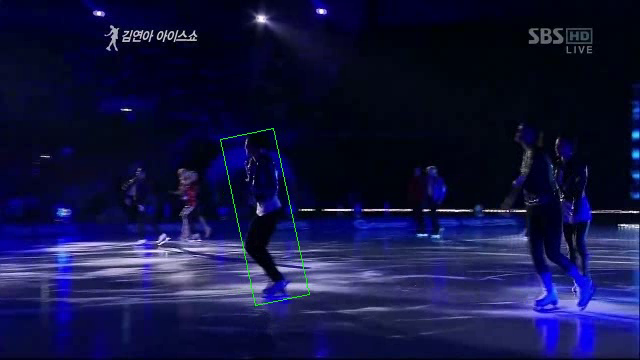}}
	           \label{fig:skating}}
\caption{A few example where the precision of the descriptors differs consistently. In sequence~(a) AKAZE the camera keeps the a constant distance from the target who is often occluded by other people. SIFT and AKAZE performs better in this scenario. In sequence~(b) the target is often blurred due to high motion, ORB and BRISK have the best precision in such as scenario. Sequence~(c) is one of the most challenging, it contains scale and light changes and motion blur. BRISK and ORB have higher accuracy however on light changes all the descriptors lose the target.}
\vspace{-3mm}
\label{fig:tracking_comparison}
\end{figure*}

Most of the feature descriptors proved to be effective for tracking purposes showing a good precision and performance. This is positive since it makes these suitable for real-time applications. Despite being assessed as somewhat weak in terms of distinctiveness, ORB and BRISK have been proven to be good for real time frameworks. 
However, the trade-off between the distinctiveness and accuracy is not easy to define.
AKAZE and SIFT have been proven to be more distinctive as descriptors but only their implementations on the GPU allow real time performances. 
The most important factors that diversify the results are the characteristics of the video sequences: object transformations, in particular scale, object appearance, light condition and motion blur. AKAZE and SIFT have shown to be more distinctive and effective when the video does not present drastic movements. AKAZE seems to be particularly sensitive to change in scale or blurring, on the other hand it has higher performance on low textured objects and people, still in sequences where the tracked object keep a constant distance from the camera as in Fig.~\ref{fig:tracking_comparison}.
Moreover, we noticed that the matching rate of all features drop consistently upon fast movements of the camera as we discussed in \cite{pieropan15}. However, the performance of weak feature descriptors such as BRISK or ORB seems to be less sensitive to this kind of noise as in Fig.~\ref{fig:tracking_comparison}. 

\begin{figure}[!htb]
	\vspace{2mm}
\centerline{%
	\subfigure[scale]{\includegraphics[width=0.33\linewidth]{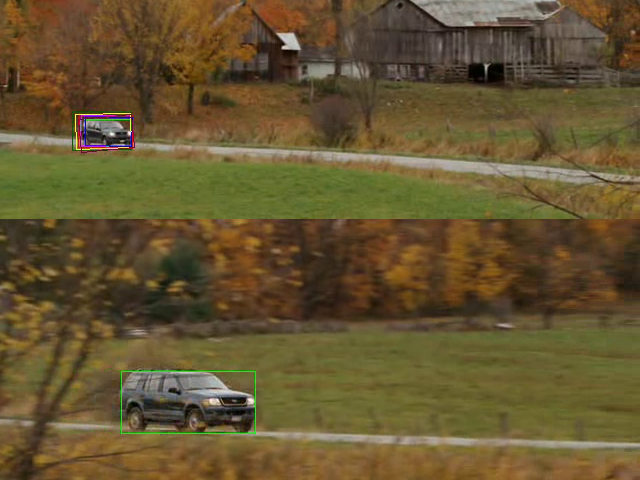}\label{fig:tra}}
	\subfigure[light]{\includegraphics[width=0.33\linewidth]{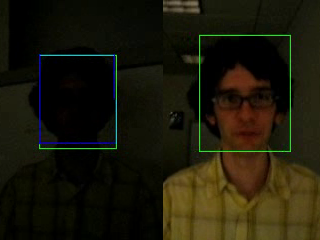}\label{fig:trb}}
	\subfigure[multi-instance]{\includegraphics[width=0.33\linewidth]{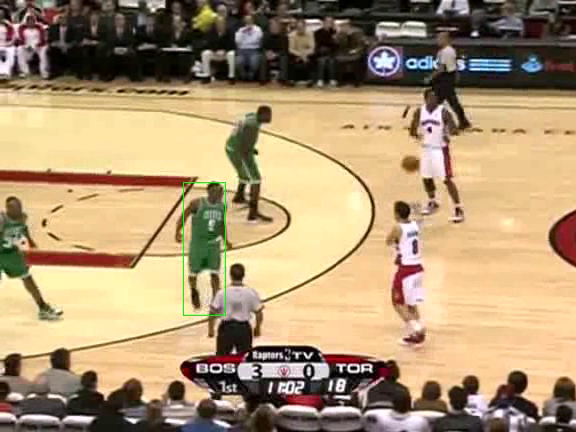}\label{fig:trc}}}
	\vspace{-2mm}
\caption{Examples showing the main problems that feature descriptors cannot address. }
\label{fig:tracking_results_scale}
\end{figure}

There are still some issue that need to be addressed in relation to achieving a robust tracking system. First, even if many descriptors are invariant to scale or rotation, we noticed that this does
not hold for drastic changes like in Fig.~\ref{fig:tra}. One possible solution is to extract features from appearances of the object generated through synthetic transformations. It has been shown by Morel \cite{morel2009} that this technique improves the matching performance of SIFT descriptor.  
Second, feature descriptors are sensitive to light conditions, see Fig.~\ref{fig:trb}. This is particularly relevant for robotics applications since the interaction of a robotic platform with a target object may occlude the light source. Third, the common matching approach to detect an object or compute the transformation between images \cite{mikolajczyk05} does not work in the presence of multiple targets with similar appearance. This is the case shown in Fig.\ref{fig:trc} where more players have the same outfit.

\section{Conclusion}

We performed an evaluation of the most common feature descriptors for the purpose of tracking by detection. Our experiments have shown that most of the feature descriptors have comparable results on a dataset presenting challenges like motion blur, occlusion, scale, rotation and light changes. AKAZE and SIFT have proven to be more distinctive with small object transformation and motion blur. AKAZE seems to be particularly sensitive to scale changes. On the other hand ORB and BRISK have better performance on sequences with significant motion blur. Given the growing interested in AKAZE descriptor, we implemented the method using CUDA and evaluated the impact of a non-linear filtering techniques, not particularly suited for a GPU's architecture, on the performance of the AKAZE feature descriptor. Our results have shown that benefit of the extracting and matching a binary descriptor compensates the more demanding detection step achieving similar performances of the implementation of SIFT on the GPU. All the code to perform the benchmark, the dataset and our implementation of AKAZE will be publicly available in order to ease researches in this area.

\bibliographystyle{unsrt}
\bibliography{main}

\end{document}